\documentclass{article}
\usepackage[utf8]{inputenc}
\usepackage[T1]{fontenc}
\usepackage[english]{babel}
\usepackage[scaled=.95]{zi4} 
\usepackage{microtype}
\usepackage{graphicx}
\usepackage{calc}
\usepackage{color}
\usepackage{float}
\usepackage{xcolor}
\usepackage{xspace}
\usepackage{scalefnt}
\usepackage{amsmath}
\usepackage[inline]{enumitem}
\usepackage{multirow}
\usepackage{listings}
\usepackage{epsfig}
\usepackage{standalone}
\usepackage[nolist]{acronym}
\usepackage[ruled,linesnumbered]{algorithm2e}

\SetCommentSty{mycommfont}
\usepackage{algorithmic}
\usepackage{pgf}
\usepackage{pgfplots}
\usepackage{pgfplotstable}
\usepackage{tikz}
\usepackage[caption=false]{subfig}
\usepackage{url}
\usepackage{todonotes}
\usepackage{acronym}
\usepackage[defaultlines=2]{nowidow}
\usepackage[%
    style=ieee,%
    backend=biber,%
    maxnames=8,%
    minnames=1,%
    mincitenames=1,%
    maxcitenames=2,%
    language=american,%
    firstinits=true,%
    doi=false,%
    isbn=false]{biblatex}
\usepackage{lipsum}
\usetikzlibrary{
    shapes, positioning, calc, backgrounds,
    shapes.arrows, arrows,
    arrows.meta,
    decorations.pathmorphing,
    decorations.markings,
}
\usepgfplotslibrary{groupplots}

\ifx\print\empty%
  \def\final{}%
\fi

\usepackage[preprint, nonatbib]{head/neurips_2020}
\usepackage[utf8]{inputenc} 
\usepackage[T1]{fontenc}    
\usepackage{hyperref}       
\usepackage{cleveref}
\usepackage{url}            
\usepackage{amsfonts}       
\usepackage{nicefrac}       
\usepackage{microtype}      
\usepackage{algorithmic}
\usepackage{environ}
\usepackage{tikzscale}
\usepackage[autostyle]{csquotes}
\newacro{AST}{Abstract Syntax Tree}
\newacro{CV}{Computer Vision}
\newacro{CNN}{Convolutional Neural Network}
\newacro{CAM}{Class Activation Map}
\newacro{SDFG}{Synchronous Data Flow Graph}
\newacro{AI}{Artificial Intelligence}
\newacro{DNN}{Deep Neural Network}
\newacro{FPGA}{Field Programmable Gate Arrays}
\newacro{CNN}{Convolutional Neural Network}
\newacro{IWMSE}{Importance-Weighted Mean Square Error}
\newacro{LRP}{Layer-wise Relevance Propagation}
\newacro{MSE}{Mean Square Error}
\newacro{MAC}{Multiply-Accumulate Operation}
\newacro{NPs}{Number of Parameters}
\newacro{GPU}{Graphics Processing Unit}
\newacro{ILSVRC2012}{Large Scale Visual Recognition Challenge 2012}
\newacro{DeepLIFT}{Deep Learning Important Features}
\newacro{SHAP}{Shapley Additive Explanations}

\newcommand{\etal}{\textit{et al.~}}

\title{Utilizing Explainable AI for Quantization and Pruning of Deep Neural Networks}

\author{%
  Muhammad Sabih, Frank Hannig and J{\"u}rgen Teich
  \\
  Hardware/Software Co-Design, Department of Computer Science\\
  Friedrich-Alexander University Erlangen-N\"urnberg (FAU)\\
  Cauerstr.~11, 91058 Erlangen, Germany\\
  \texttt{\{muhammad.sabih, frank.hannig, juergen.teich\}@fau.de} \\
}

\begin{document}
\maketitle
\begin{abstract}
For many applications, utilizing \acp{DNN} requires their implementation on a target architecture in an optimized manner concerning energy consumption, memory requirement, throughput, etc.
\ac{DNN} compression is used to reduce the memory footprint and complexity of a \ac{DNN} before its deployment on hardware. Recent efforts to understand and explain AI (Artificial Intelligence) methods have led to a new research area, termed as explainable AI. Explainable AI methods allow us to better understand the inner working of \acp{DNN}, such as the importance of different neurons and features.
The concepts from explainable AI provide an opportunity to improve \ac{DNN} compression methods such as quantization and pruning in several ways that have not been sufficiently explored so far. In this paper, we utilize explainable AI methods: mainly DeepLIFT method. We use DeepLIFT for (1) pruning of \acp{DNN}; this includes structured and unstructured pruning of \ac{CNN} filters pruning as well as pruning weights of fully connected layers, (2) non-uniform quantization of \ac{DNN} weights using clustering algorithm; this is also referred to as Weight Sharing, and (3) integer-based mixed-precision quantization; this is where each layer of a \ac{DNN} may use a different number of integer bits.
We use typical image classification datasets with common deep learning image classification models for evaluation.
In all these three cases, we demonstrate significant improvements as well as new insights and opportunities from the use of explainable AI in \ac{DNN} compression.
\end{abstract}

\section{Introduction}
\acp{DNN} have achieved rapid success in many image processing applications, including image classification~\cite{AlexNet,Resnet}, image segmentation~\cite{Unet,deeplab}, object detection~\cite{MaskRCNN}, etc. The key ingredients in this success of \acp{DNN} have been the usage of deeper networks and a large amount of training data. However, as the network gets deeper, the model complexity also increases rapidly. 
The training of \acp{DNN} can be carried out on high-performance clusters with \ac{GPU} acceleration; however, for implementing these networks on hardware, the complexity of the \acp{DNN} needs to be reduced. This includes decreasing the memory requirements, energy consumption, latency or throughput of the implementation of \acp{DNN} on the hardware.
Pruning and quantization are two of the main techniques that are utilized for compression and complexity-reduction of \acp{DNN}. Our work is motivated by the increasing requirement of taking the \acp{DNN} to the hardware as well as the progress in understanding and explaining AI. \ac{DNN} pruning and quantization have been a popular topic for research recently, and a number of works have been published in this area.

Few of the challenges in this area of research are: lack of standard metrics, datasets and benchmarks and the lack of explainable algorithms that can  achieve state-of-the-art or close to the state-of-the-art for a broad range of problems: we typically have one approach performing better for one application and another approach performing better for another application.
With this research work, we aim to not only use explainable AI for better performing pruning and quantization algorithms but also to address a broad range of cases with a unified approach: this includes structured and unstructured pruning, \ac{CNN} filter pruning and neuron pruning, mixed-precision integer-based quantization and weight sharing quantization.
Additionally, our proposed approach can optimize for \acp{MAC}, \ac{NPs}, or both.
\subsection{Contributions}
Our novel contributions in this paper can be summarized as follows:
\begin{enumerate}
  \itemsep0em 

\item Utilizing explainable AI, in particular, DeepLIFT, for obtaining the importances of neurons for the pruning and quantization of \acp{DNN}.
\item Proposing a method to calculate sensitivities of layers for quantization and pruning.
\item Using (1) and (2) with iterative pruning algorithm for structured and unstructured pruning can optimize \acp{MAC}, \ac{NPs}, or both.
\item Using (1) with a weighted MSE criteria to obtain better clustering of weights.
\item Using (1) and (2) and (4) to obtain bit-widths for mixed-precision integer quantization.
\item Obtaining state-of-the-art or competitive results for these methods.
\end{enumerate}

\section{Explainable AI}
Obtaining the importances of layers, neurons and filters is a common challenge in quantization and pruning of \acp{DNN}.
This can be addressed using explainable AI. 
Among the different types of explainable AI algorithms, we are most interested in the \emph{saliency} methods.
They explain an algorithm's decision by assigning values that reflect the importance of input
components in their contribution to that decision~\cite{SurveyExplainableAI}. These methods such as \ac{CAM}~\cite{CAM}, \ac{LRP}~\cite{LRPPruning}, \ac{DeepLIFT}~\cite{DeepLIFT}, Conductance~\cite{Conductance}, etc., are types of saliency methods that can be used to obtain the importance measures. 

\subsection{\ac{DeepLIFT}}
In~\cite{DeepLIFT}, the authors describe the algorithm as a method for decomposing the output prediction of a neural network on a specific input by back-propagating the contributions of all neurons in the network to every feature of the input.
DeepLIFT compares the activation of each neuron to its \emph{reference activation} and assigns contribution scores according to the difference.
The choice of reference activation is important for the algorithm's outcome, and it often requires domain-specific knowledge.
In order to specify a reference activation, we must understand the intuition behind the DeepLIFT algorithm. It compares the effect of the features to a baseline of what the model would predict when it can not see the features. Therefore, a good reference activation for MNIST~\cite{MNIST} is an all-black image.
For more complex datasets such as CIFAR10, CIFAR100, and \ac{ILSVRC2012}, different possibilities for reference activations are mean-image of the training set, mean-image of the training set with random noise, all zero image, noisy versions of input images, etc.

Mathematically, the DeepLIFT algorithm is formulated by the authors as follows. Let $t$ represent some target output neuron of interest and let $x_1, x_2,\dots, x_n$ represent some neurons in some intermediate layer or set of layers that are necessary and sufficient to compute $t$. Let $t^0$
represent the reference activation of $t$. The quantity $\Delta t$ is defined as the difference-from-reference, that is $\Delta t = t - t^0$. DeepLIFT assigns contribution scores $C_{\Delta x_i \Delta t}$ to $\Delta x_i$ s.t.:
\begin{equation}
\sum^{n}_{i=1}{C_{\Delta x_i \Delta t}} = \Delta t
\end{equation}
$C_{\Delta x_i \Delta t}$ can be thought of as the amount of difference-from-reference int $t$ that is attributed to the difference-from-reference of $x_i$. $\Delta t$ is the DeepLIFT score, which can also be represented as follows.

\begin{equation}
\mathrm{DL(t, x)} = \sum^{n}_{i=1}{C_{\Delta x_i \Delta t}}
\end{equation}

Our choice for using the DeepLIFT is because it solves the problem of saturation with the gradient methods such as \ac{LRP}.
Conductance is a closely related algorithm that gives similar results.
However, DeepLIFT is a more computationally efficient method.
The draw-back of DeepLIFT and Conductance is that they need a baseline or a reference input, against which the hidden representation of the \ac{DNN} is compared. To obtain this reference or baseline, one needs domain-specific knowledge.
In their work, the authors suggested using three types of references: (1) a blank image with all zero pixels, (2) an image that is the mean of the training images in the datasets, and (3) a blurred version of the input image.

\section{DNN Pruning}
Pruning refers to removing the undesired parameters of a \ac{DNN} that have little influence on the output of the neural network~\cite{PruningIntro1, PruningIntro2}.
This leads to fewer \acp{MAC} operations and fewer \ac{NPs}.

Pruning can be \emph{structured} or \emph{unstructured}. In structured pruning, the filters and weights are removed by removing all their input and output connections, and this means that no extra compilation effort or hardware optimization is required to get the gain on hardware in compression or the inference efficiency.
In unstructured pruning, the unimportant filters or weights are set to zero, a compiler utilizes these zeros to get a gain in compression or efficiency in inference. Utilizing the unstructured pruning at the hardware has additional cost in terms of compilation effort or computational effort in order to exploit the irregular sparsity. Typically unstructured pruning provides more pruning ratios.

Pruning can be \emph{neuron pruning}, \emph{filter pruning}, \emph{weight pruning}, and \emph{layer pruning}.
In neuron pruning, individual neurons are removed, i.e., all the incoming and outgoing connections to a neuron are also removed~\cite{NeuronPruning}. In filter pruning, \ac{CNN} filters are removed~\cite{CNNFilterPruning}. In layer pruning, some of the layers can also be pruned~\cite{CVPRLayerPruning}. Weight pruning is used synonymously for unstructured pruning, where the redundant weights are set to zero.


The two fundamental objectives for pruning the model are (1) reducing the memory by lowering the \ac{NPs} and reducing the latency and energy consumption of the hardware by reducing \acp{MAC}. These two objectives are often conflicting in a \ac{DNN} because the \acp{MAC} are concentrated in the lower layers of a typical image classification network and the number of parameters are concentrated in the higher layers of a typical image classification network.

The key challenge in the pruning of \ac{DNN} is identifying criteria by which the importance of parameters can be measured. This importance can be of two types: one is the importance of parameters based on their contribution or effect on the output of the network: this is typically measured by the degradation in accuracy of the \ac{DNN} output after pruning that parameter, the other type of importance is the reduction in memory or number of multiplications that is obtained after removing the parameters.
The effect that a \ac{DNN} parameter has on \acp{MAC} or memory is typically constant for each layer.
However, the effect that a parameter has on the output of the network varies considerably within the layer.

\subsection{Related work and State-of-the-art}

Some of the examples of commonly used criteria for pruning are \( \ell^1 \) norm and \( \ell^2 \) norm of the weights.
The closest work to ours for \ac{CNN} filter pruning is~\cite{LRPPruning}, where authors used explainable AI method: \ac{LRP}, to calculate the importance of features. In their work, authors provide useful insights to utilize pruning to focus on some of the output classes.
\ac{LRP} is a gradient-based method and suffers from similar limitations against saturation of activations, as shown by Shrikumar~\etal~\cite{DeepLIFT}.
In~\cite{HessianPruning}, a second-order Taylor expansion
based on the Hessian matrix of the loss function was used to select parameters for pruning, however computational cost of calculating the exact Hessian is prohibitive and approximations lead to incorrect importance measures.

The state-of-the-art for \acp{DNN} is challenging to identify. There are various reasons for this such as: (1) this is a relatively new research area and there are a lack of standardized benchmarks, (2) there is a difference in implementation of networks on different deep learning platforms, (3) the choice of the un-pruned reference network has different effects on the results, (4) the training recipes and training budgets considerably vary, (5) different types of metrics and combinations of networks and datasets are used by different authors, etc.
In the work titled \citetitle{PruningSOTA}~\cite{PruningSOTA}, the authors survey 81 papers and summarize the state-of-the-art results related to pruning of \acp{DNN} and also suggested standard practices for benchmarking \ac{DNN} pruning. 


\subsection{DeepLIFT-based local pruning}
We propose a DeepLIFT-based local pruning approach that prunes \ac{CNN} filters or neurons, locally in a layer according to the importances calculated from DeepLIFT method. In local pruning, we prune a layer or a group of layers with the same specific amount and optionally fine-tune it.

\begin{figure}
\centering
  \includegraphics[width=\linewidth]{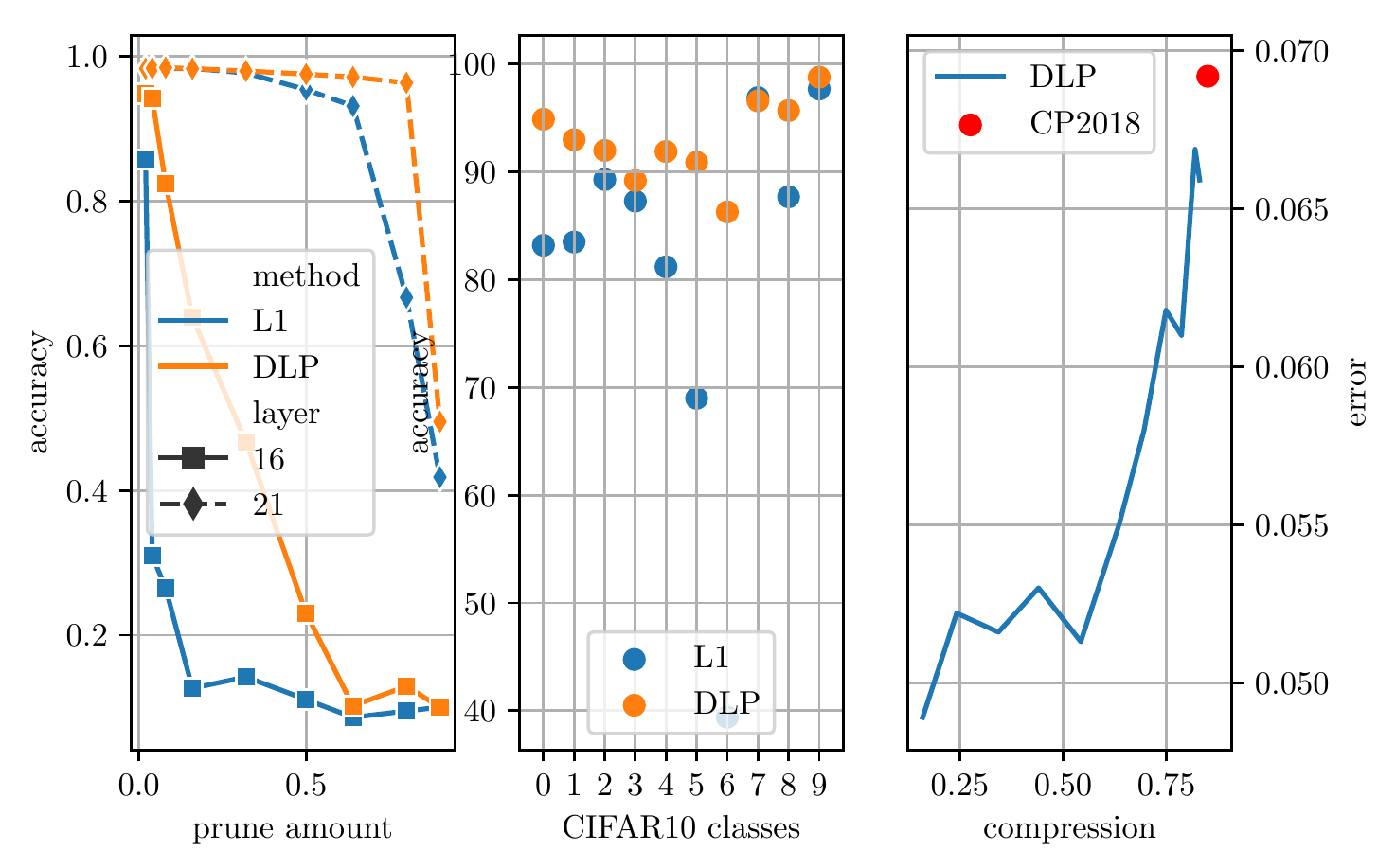}
  \caption{In the first figure from the left, we prune the last layer (21) of ResNet20 (after counting down-sampling operations as layers) and layers 16 to 21 with varying pruning amounts and plot testing accuracy on the y-axis. We do not use any fine-tuning; this helps us understand how DeepLIFT Pruning (DLP) criteria are better than the baseline of \( \ell^1 \) norm. In the second figure from the left, we prune the last layer of ResNet20 but evaluate each CIFAR10 class separately for a pruning amount of 0.5. This explains to us how, for some classes, the lower magnitude of weights corresponds to lower importance, but for others, it does not. DeepLIFT-based pruning criteria give us the degree of freedom to prioritize some output classes over others. In the third figure from the left, we compress the ResNet56 model on CIFAR10 using our global unstructured pruning method. It is also abbreviated as DLP for varying compression ratios and testing accuracies, along with the state-of-the-art point from CP2018~\cite{UnstructSOTA}. Our error is lower, but our compression rate is also slightly lesser than the state-of-the-art.}
  \label{fig::AllPrun}
\end{figure}

This type of pruning is useful, if the goal is to reduce the memory requirement or \acp{MAC} of a layer or a group of layers to a certain amount, for example, if on a \ac{FPGA}, a certain layer cannot fit on the memory or if the latency of a particular layer is too much, then pruning of unimportant parameters is needed. Local pruning without fine-tuning or retraining is also useful to evaluate the performance of the ranking criteria for pruning because it gives a measure of how well the ranking criteria identifies the redundant features. With training or fine-tuning, the performance of a pruning method can be attributed to a good training or fine-tuning recipe rather than the pruning criteria.
We compare the results of our local pruning approach by comparing it with 
the pruning based on \( \ell^1 \) norm of the weights, in order to have an initial assessment of the performance of our proposed method. The results are shown in \Cref{fig::AllPrun}.

Another interesting application of DeepLIFT-based pruning criteria is to prune such that the accuracy of a subset of image classes is given more preference than the others. For our work, we do not use this degree of freedom.

We describe our algorithm as follows in \Cref{alg::locprun}
\begin{algorithm}
    \small
    \caption{DeepLIFT-based local pruning}\label{alg::locprun}
    \hspace*{\algorithmicindent} \textbf{Input} layers\_to\_prune, prune\_amount, trained\_model \\
    \hspace*{\algorithmicindent} \textbf{Output} pruned\_model
    \begin{algorithmic}[1]
    \FOR{each\_layer in layers\_to\_prune}
        \STATE calculate the \( \ell^1 \) norm of DeepLIFT importances.
        \STATE sort the importance.
        \STATE prune the lowest weights specified by prune\_amount.
    \ENDFOR
    \RETURN{} pruned\_model
\end{algorithmic}
\end{algorithm}

\subsection{DeepLIFT-based layer sensitivity analysis}

The importance measures obtained from DeepLIFT explain how much one neuron or filter effects the output of the layer, these importances do not tell the global importance.

\begin{figure}
\centering
  \includegraphics[height=4.8cm]{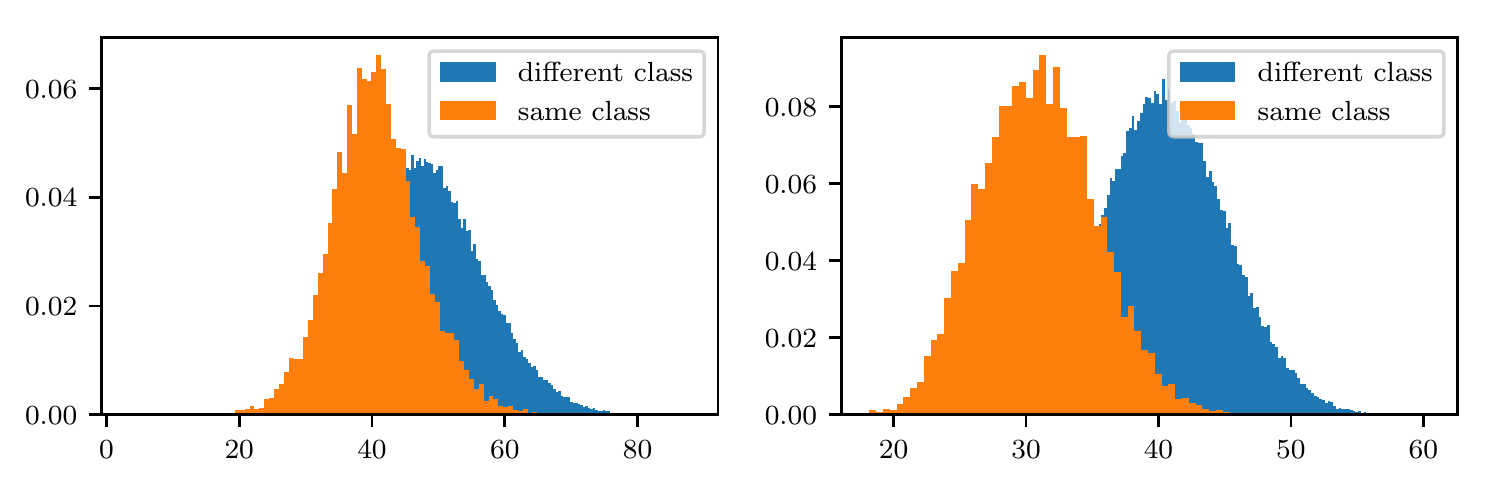}
  \caption{In the figure, we show how the distributions of same-class distances and different-class distances vary from layer to layer. The figure on the left shows how distributions of pairwise distance at the last convolutional layer look like after introducing distortion in the first layer of ResNet20. The figure on the right shows the same for the second last convolutional layer of ResNet20.
}
  \label{fig::sens}
\end{figure}

This importance can be obtained by scaling the local importance measures with the sensitivities of each layer.
\textcolor{black}{In order to obtain these sensitivities, we propose a \emph{DeepLIFT-based sensitivity} analysis of the \ac{DNN}.}
\acp{DNN} can be modeled as a Markov chain~\cite{IB}, such that the layered structure of a network generates a successive Markov chain of intermediate representations, which together form (approximate) sufficient statistic.
We are interested in measuring how much distortion in a layer (whether by pruning or quantization) results in how much degradation of the network's output.
We use the interpretation of a \ac{DNN} as a Markov chain to say that each layer in a \ac{DNN} takes the representation from the previous layer and adds to the clustering ability of the \ac{DNN} such that in an intermediate representation, the two images belonging to the same class are closer to each other and farther away from the images of the different classes (for example, a cat image and a dog image).
This means we can measure the pairwise distances from the important parts of the intermediate representation of image classes to determine how good is the separability of the image classes at a layer. So we obtain the sensitivities for each layer by introducing distortion into that layer by quantization or pruning and then measuring the separability of classes in terms of difference between the distribution of same-class distances and the distribution of different-class distances. 
In~\Cref{fig::sens}, the separability of first layer and last layer of a ResNet-20 model on CIFAR10 is shown in terms of distributions of same class distances and different class distances. 

\subsection{DeepLIFT-based global pruning}
Often, the goal of \ac{DNN} pruning is not just to prune a specific layer by a specified number, but the goal is to globally prune the \ac{DNN} the filters of a \ac{CNN} and neurons of fully-connected layer, in order to minimize the \ac{NPs} (memory), \acp{MAC} (latency) or both.
To obtain the global importance, a typical approach is to rank the filters or neurons according to criteria, prune a fraction of these filters, tune the \ac{DNN} and repeat the process until a certain requirement is met.
We take a slightly different approach, we first calculate the number of filters or neurons that are need to be pruned in each layer, this number is calculated from the sensitivities of the layers in addition to  an objective such as the number of \acp{MAC} or \ac{NPs} in the layer, after this, the importance measures are used within the layers to prune the least important filters or neurons. 
One can use this algorithm to prune the \ac{DNN} to optimize for \acp{MAC}, \ac{NPs}, or both. Our algorithm is described in \Cref{alg::glob}.
\begin{algorithm}
    \small
    \caption{DeepLIFT-based global pruning}\label{alg::glob}
    \hspace*{\algorithmicindent} \textbf{Input}  trained\_model, \\
    \hspace*{\algorithmicindent} \textbf{Output} pruned\_model
    \begin{algorithmic}[1]
    \WHILE{error < maximum error and iterations < maximum iterations }
    \STATE get layer sensitivities 
    \IF {optimization criteria == \acp{MAC}}
        \STATE get load of layers by \acp{MAC}.
    \ELSIF {optimization criteria == \ac{NPs}}
        \STATE get load of layers  by \ac{NPs}.
    \ENDIF
     \STATE multiply sensitivities of layers with their load to get pruning number per layer.

    \FOR{each\_layer in model}
        \STATE calculate \( \ell^1 \) norm of DeepLIFT importances for the layer.
        \STATE sort the importances.
        \STATE prune the number of filters or neurons for this layer.
    \ENDFOR
    \ENDWHILE
    \RETURN{} pruned\_model
\end{algorithmic}
\end{algorithm}

\subsubsection{Evaluation}

\begin{figure}
\centering
  \includegraphics[height=6cm]{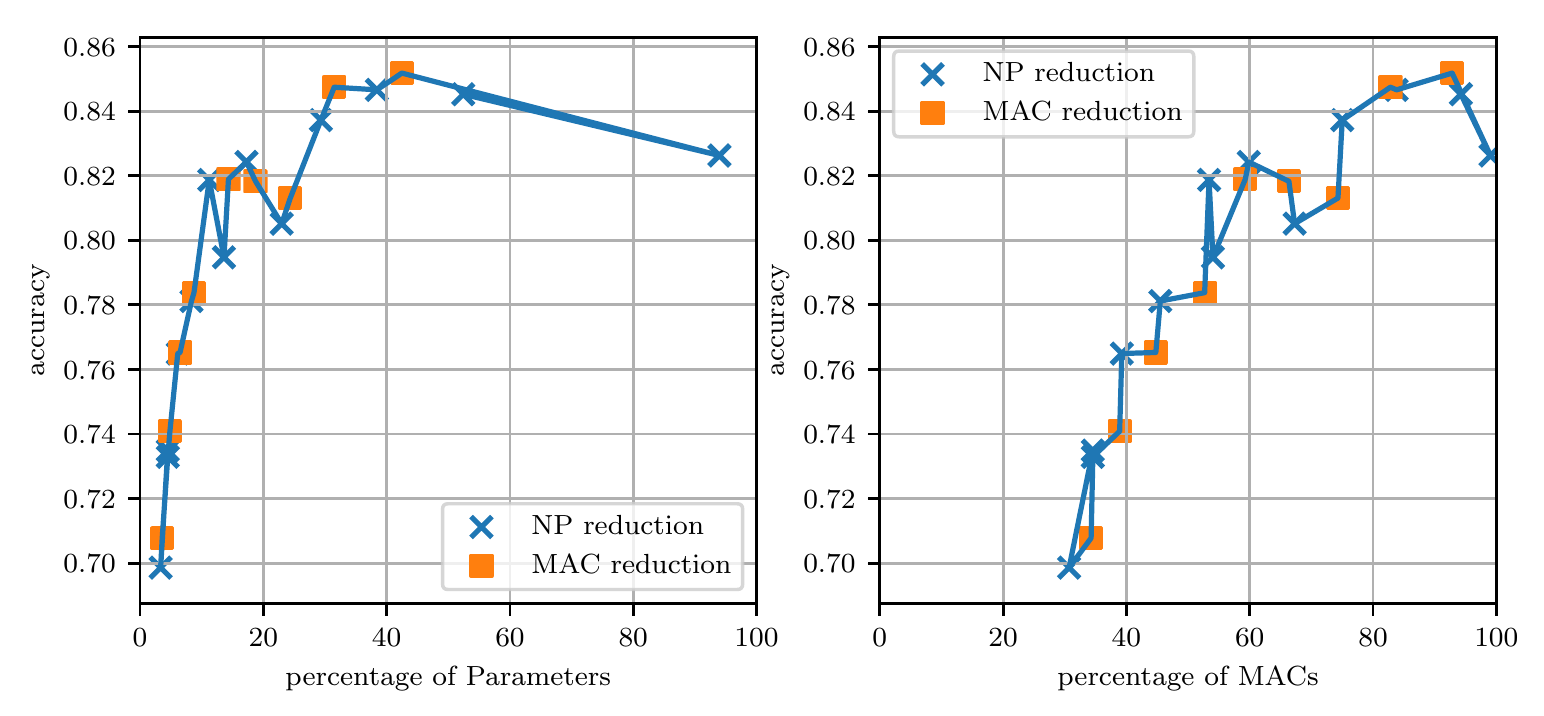}
  \caption{This figure shows the results from DeepLIFT global pruning method. The figure on the left compares the reduction in the number of parameters with the degradation in accuracy as the pruning algorithm iterates. The iterations start from right. The figure on the right compares the number of \acp{MAC} with the degradation in accuracy. One pruning iteration reduces \acp{MAC}, and the other reduces \ac{NPs}. Both objectives, i.e., \ac{MAC} reduction and NP reduction, are supporting each other in the start because the reduction in \ac{NPs} of an over-parameterized network helps reduce \acp{MAC} without loss in accuracy. However, later on, these two objectives conflict with each other.}
  \label{fig::StrucPrun}
\end{figure}

We choose the Taylor approximation of the Hessian approach~\cite{HessianPruning} and the \ac{LRP}~\cite{LRPPruning} approach as our comparison methods. One datasets and model pair used in these two papers is AlexNet~\cite{AlexNet} with Oxford Flowers Dataset~\cite{OxfordFlowers}. This is a 102 category-dataset with around 2000 images for training and around 8000 for testing. The resolution of images roughly matches that of ImageNet~\cite{ImageNet}, so this is a good lighter alternative to ImageNet for benchmarking.
In both the LRP-based approach and the Taylor approximation approach, the goal is to prune the \ac{CNN} filters for optimal \acp{MAC} and \ac{NPs}. In our case, we do not restrict ourselves to \ac{CNN} filters alone.
However, our goal is also to prune AlexNet for optimal \acp{MAC} and \ac{NPs}.
We start with AlexNet pre-trained on ImageNet and perform transfer learning to obtain an initial accuracy of 81.9\% on the testing dataset.
We use our DeepLIFT global pruning algorithm by using one pruning round for optimizing \acp{MAC} and one pruning round for optimizing \ac{NPs}. In each round, 10 to 15 percent of \acp{MAC} or \ac{NPs} need to be reduced. Between each pruning round, we fine-tune or retrain the AlexNet weights by training for 30 epochs with learning rate varying between 0.01 and 0.001 and using SGD~\cite{SGD} optimization. For DeepLIFT, we use blurred images corresponding to the input image as a reference
These blurred images are obtained from Gaussian blurring, and the DeepLIFT importances are obtained from 512 training images. The results of our experiment are shown in \Cref{fig::StrucPrun}.

In terms of \acp{MAC}, our approach can keep up with the Taylor approach in terms of accuracy up to the \ac{MAC} reduction of around 70\%; these are 237 Million \acp{MAC} (MMACs) or 0.47 GFLOPs in absolute terms and roughly 3x less than the original \acp{MAC}. The accuracy of the model at this point on the test dataset is 0.70.
This is despite the fact, that in terms of \ac{NPs}, our model is orders of magnitude more compressed, because in the Taylor approach and the LRP approach, only \ac{CNN} filter maps are considered for pruning. However, we argue that in terms of the end goal, we are mostly interested in the overall reduction in \acp{MAC} and \ac{NPs}, rather than specifically pruning either the \acp{CNN} or the fully connected layers. Our final results are shown in \Cref{tab::StrucPrun}

\begin{table}[]
\centering
\caption{Our most pruned model contains 30x less parameters (0.30\%) and 3x less GFLOPs (0.30\%) than the un-pruned model.}
\label{tab::StrucPrun}
\begin{tabular}{|l|l|l|l|l|}
\hline
                       & Accuracy & Reference Accuracy & \ac{NPs} & \acp{MAC} \\ \hline
AlexNet on Flowers 102 & 0.698 & 0.819 & 2.03 million & 0.47 GLOPs \\ \hline
\end{tabular}
\end{table}
\subsection{Unstructured Pruning}
For the unstructured pruning, we use the same approach as in global pruning, except that in the case of unstructured pruning, we can prune the individual weights in the \ac{CNN} filter's kernel. The DeepLIFT importance measures can only be calculated at the granularity of a \ac{CNN} channel, so we multiply the DeepLIFT importance measures with the \( \ell^1 \) norm of the weights, with this modification, we can use the DeepLIFT-based global pruning algorithm for unstructured pruning as well.

For evaluating our unstructured pruning approach, we refer to the shrink-bench~\cite{PruningSOTA} for the state-of-the-art. We choose ResNet56 on CIFAR10 as our competitor, where the state-of-the-art compression scores are reported by Carreira-Perpinan 2018 (CP2018)~\cite{UnstructSOTA}. We iteratively prune the model in 10 iterations, and training with SGD and varying learning rates between 0.1 and 0.001 are employed, each pruning iteration trains for 40 epochs.  
Our pruning method gives an error of 0.066 on the testing set of CIFAR10 dataset, with 83\% of the weights pruned. Our scores are well above the various baselines for CIFAR10 and Resnet56, reported in shrink-bench~\cite{PruningSOTA}. In CP2018, the reported score for ResNet56 CIFAR10 is an error of 0.0692, with 85\% of the weights pruned.
A lack of a common reference model and non-availability of a full trade-off curve makes our comparison difficult. Our trade-off curve is shown in \cref{fig::AllPrun}.

\section{DeepLIFT-based quantization for weight sharing}

Quantization for \acp{DNN} refers to approximating a neural network using reduced bit precision. This was first used in \cite{QuantizationIntro1} and \cite{QuantizationIntro2}. 
\acp{DNN} can be quantized using a clustering algorithm, whereby, all the weights that fall into a cluster share the same value. For example, 32-bit weight values can be quantized to a 2-bit value using four clusters. This is an example of non-uniform quantization (with non-uniformly separated quantization points), and the idea is also known as weight sharing~\cite{WeightSharing}. 
Han~\etal~\cite{Han2016DeepCC} used weight sharing quantization along with pruning and Huffman coding to reduce the \ac{DNN} size by 35 to 50 times.
We propose DeepLIFT Weighted Mean Squared Error (DWMSE) criteria that can be used with k-means clustering algorithm~\cite{K-means} to obtain a more efficient quantization of the \ac{DNN} weights. In the proposed scheme, each \ac{DNN} weight has an associate DeepLIFT score and a scaling factor based on the number of parameters a layer has in proportion to other layers in the whole network.

\begin{figure}
\hspace{-3mm}
  \includegraphics[height=5cm]{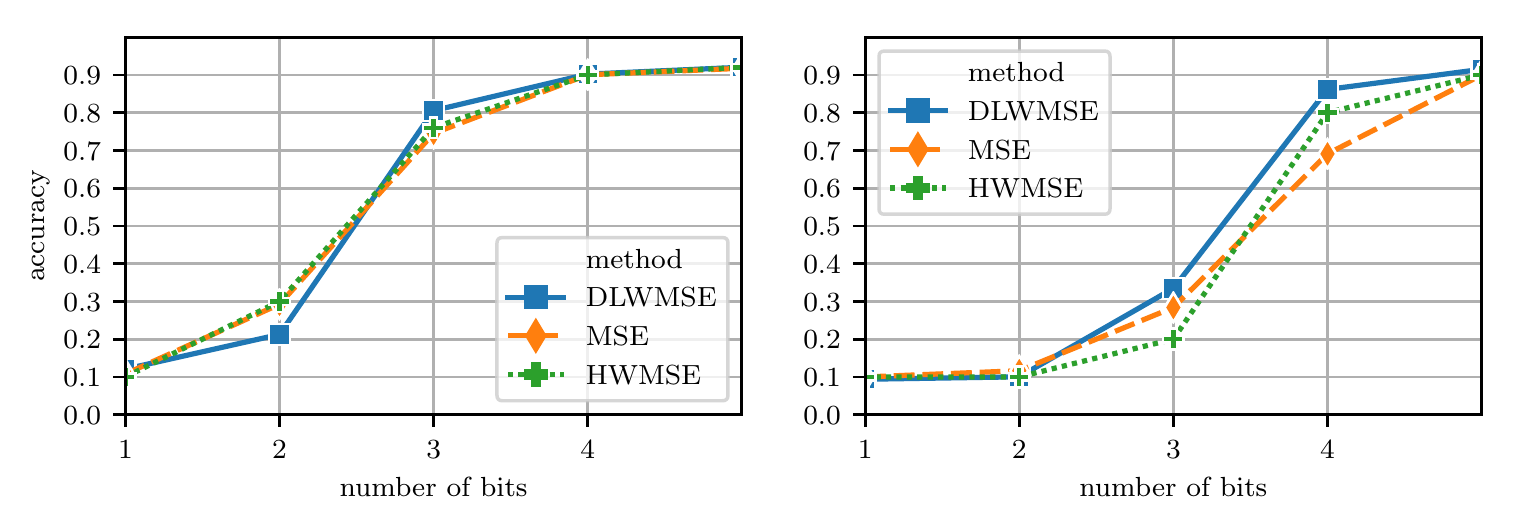}
  \caption{This figure compares Hessian Weighted MSE (HWMSE) criteria with DeepLIFT Weighted MSE (DLWMSE) and MSE. The figure on the left shows results after batch-norm calibration or fine-tuning. The figure on the right shows results without fine-tuning. The accuracies are calculated on the testing set of CIFAR 10. Our approach gives better accuracy, especially when no fine-tuning is employed.}
  \label{fig::WeightSharing}
\end{figure}

We compare our results with~\cite{HessianWeighted}; in this work, the authors used the weights from the Taylor approximation of Hessian to calculate the weighting factor of k-means.
As the authors use ResNet20 with CIFAR10 for evaluation, we use the same dataset and model pair. The plain k-means quantization with \ac{MSE} criteria is used as an additional baseline. We first quantize the weights without recalibrating the batch norm and with recalibrating the batch norm. We do not retrain our weights. In~\cite{HessianWeighted}, the authors also presented their results without fine-tuning and with fine-tuning. Their fine-tuning might also mean additional training.
They additionally utilize pruning to compress the model further.
However, we only compare our results with their quantization. These results are shown in \Cref{fig::WeightSharing}.

\section{DeepLIFT based mixed-precision integer quantization}
The \ac{DNN} can also be quantized using integer-based uniform quantization.
This speeds up inference on different kinds of hardware and also reduces the memory requirement. 

We propose DeepLIFT based search, whereby, instead of reducing the bit-precision one by one, we collectively reduce the bit-precisions in all layers, according to the sensitivities of the layers. We call this coarse search. At some point when the accuracy goes below a certain threshold, we can stop these coarse iterations and start the iterations, whereby, the bit-precision of each layer can be changed, in the order of their sensitivities. 
In~\cite{hawq}, the Hessian is used to sort the layers according to their sensitivities, and then a greedy search is performed to obtain the optimum bit-widths. Their results along with ours for ResNet20 CIFAR10 are briefly summarized as: We obtain an accuracy of 91.2\% on CIFAR10 dataset with the 79 Cumulative Bits (CB) for weights and 79 CB for activations, whereas in~\cite{hawq}, an accuracy of 92.22\% was obtained with 81 CB for weights and 88 CB for activations. For our experiment, was used five coarse iterations and a training step in between utilizing SGD with a learning rate varying from 0.01 to 0.001 for 30 epochs. We did not optimize specially for \acp{MAC} or \ac{NPs}.

\section{Hardware and software setup}
For all experiments, we used NVIDIA TITAN RTX GPU with 24 GB of memory and Intel's i7-9700 CPU. For software, we used Python~\cite{Python} programming language for all experiments, PyTorch~\cite{PyTorch} for deep learning, NumPy~\cite{numpy} for array manipulation, Matplotlib~\cite{matplotlib} for visualization, scikit-learn~\cite{scikit-learn} for k-means and pairwise distances, Captum~\cite{captum2019} for DeepLIFT and explainable AI, PyTorchCV~\cite{PyTorchCV} for \ac{DNN} implementations, NEMO~\cite{nemo} for mixed-precision quantization-aware training, ptflops~\cite{ptflops} for profiling \acp{MAC} and \ac{NPs}.

\section{Conclusion}
In conclusion, we utilize the DeepLIFT method for \ac{DNN} quantization and pruning. DeepLIFT is part of the algorithms in the explainable AI area, and it is considered to have an advantage over other methods for similar tasks. This advantage is also reflected in \ac{DNN} compression and inversely, we can utilize \ac{DNN} compression to assess the explainable AI methods. We additionally propose our novel algorithms utilizing DeepLIFT for a wide range of tasks in \ac{DNN} pruning and quantization like structured pruning, unstructured pruning, \ac{CNN} filter pruning, weights pruning, mixed precision integer quantization and weight sharing quantization, for optimizing memory as well as latency.
Besides achieving state-of-the-art or competitive results on most problems, this work also provides a unified approach for a diverse set of compression problems, which can be very useful for its application for real-world problems.

\section*{Acknowledgments}
This work was partially funded by the German Federal Ministry of Education and Research (BMBF) within the project KISS (01IS19070B).




\printbibliography
\end{document}